\pdfoutput=1

\documentclass[11pt]{article}

\usepackage[preprint]{acl}

\usepackage{times}
\usepackage{latexsym}
\usepackage{graphicx}
\usepackage{booktabs} %
\usepackage{multirow} %
\usepackage{array}    %
\usepackage{amsmath}
\usepackage{graphicx}
\usepackage{array}
\usepackage{booktabs}
\usepackage{siunitx}
\usepackage{adjustbox}

\usepackage{xcolor,colortbl}
\definecolor{green}{rgb}{0.1,0.1,0.1}

\usepackage[T1]{fontenc}

\usepackage[utf8]{inputenc}

\usepackage{microtype}

\usepackage{inconsolata}

\usepackage[textsize=scriptsize]{todonotes}

\usepackage{graphicx}
\usepackage{subcaption}

\newcommand{\tool}{{\sc QSalience}}
\newcommand{\data}{{\sc QSalience}-\emph{data}}

\usepackage{xcolor}
\newcommand{\highlight}[1]{{\color{magenta}\{\{#1\}\}}}

\usepackage[skins,breakable]{tcolorbox}
\usepackage{tikz}
\usetikzlibrary{tikzmark,decorations.pathreplacing,calc}

\definecolor{light-purple}{RGB}{151,156,171}
\definecolor{blue-color}{RGB}{40,166,189}
\definecolor{pink-color}{RGB}{237,46,104} 
\definecolor{dark-grey-color}{RGB}{79,91,102}
\definecolor{darkbyzantium}{rgb}{0.36, 0.22, 0.33}
\definecolor{bluebell}{rgb}{0.64, 0.64, 0.82}
\definecolor{airforceblue}{rgb}{0.36, 0.54, 0.66}

\newcommand{\promptsubsection}[1]{
\setlength{\parskip}{6pt} \noindent\textbf{{#1}:}
}

\newtcolorbox[list inside=prompt,auto counter,number within=section]{prompt}[1][]{
    colbacktitle=airforceblue,
    colframe=airforceblue,
    fontupper=\footnotesize,
    boxsep=5pt,
    left=0pt,
    right=0pt,
    top=0pt,
    bottom=0pt,
    boxrule=1pt,
    enhanced, 
    breakable,
    skin first=enhanced,
    skin middle=enhanced,
    skin last=enhanced,
    #1,
}

\title{Which questions should I answer?\\
Salience Prediction of Inquisitive Questions}

\author{Yating Wu*$^{1}$,
Ritika Mangla*$^{2}$,
Alexandros G. Dimakis $^{1, 4}$,
Greg Durrett$^{2}$,
Junyi Jessy Li$^{3}$
\\
$^1$Electrical and Computer Engineering, $^2$Computer Science, $^3$Linguistics \\ The University of Texas at Austin \\$^4$BespokeLabs.ai \\
{\tt \{yating.wu, jessy\}@utexas.edu}
}

\begin{document}
\maketitle
\begin{abstract}
Inquisitive questions --- open-ended, curiosity-driven questions people ask as they read --- are an integral part of discourse processing~\cite{van1995discourse,onea2016potential,kehler-rohde-2017-evaluating} and comprehension~\cite{prince2004does}. Recent work in NLP has taken advantage of question generation capabilities of LLMs to enhance a wide range of applications. But the space of inquisitive questions is vast: many potential questions can be evoked from a given context.
So which of those should be prioritized to find answers? Linguistic theories, unfortunately, have not yet provided an answer.
This paper presents \tool, a salience predictor of inquisitive questions. \tool{} is instruction-tuned over our dataset of linguist-annotated salience scores of 1,766 (context, question) pairs. A question scores high on salience if 
answering it would greatly enhance the understanding of the text~\cite{van2003questioning}. We show that highly salient questions are empirically more likely to be answered in the same article, bridging potential questions~\cite{onea2016potential} with Questions Under Discussion~\cite{roberts2012information}. We further validate our findings by showing that answering salient questions is an indicator of summarization quality in news.
\let\thefootnote\relax\footnotetext{*Yating and Ritika contributed equally.}
\end{abstract}

\section{Introduction}
Asking questions is the natural language manifestation of human inquisitiveness: we insist on getting answers for what we are curious about since childhood~\cite{chouinard2007children}. Acquired strategies of question generation have a profound impact on education~\cite{davey1986effects,prince2004does}. In linguistics, both theoretical and psycholinguistic work argued that readers generate inquisitive questions, seeking information in a conversation or as they read \cite{van1995discourse,ginzburg1996dynamics,onea2016potential,kehler-rohde-2017-evaluating}.
In NLP, pre-trained models have enabled the generation of these \emph{open-ended, curiosity-driven, information-seeking} questions, leading to a flourish of recent work: identifying information loss between two texts~\cite{trienes2024infolossqa,cole-etal-2023-diffqg}, analyzing the diversity of news perspectives~\cite{laban-etal-2022-discord}, generating elaborations or explanations~\cite{wu-etal-2023-elaborative,fok2023qlarify}, evaluating summaries~\cite{pratapa-etal-2023-background}, asking follow-up questions~\cite{meng-etal-2023-followupqg}, decontextualization~\cite{newman-etal-2023-question}, and planning~\cite{narayan-etal-2023-conditional}.

However, the space of possible inquisitive questions is vast. Prior work~\cite{ko-etal-2020-inquisitive,westera-etal-2020-ted} showed that many distinct questions can be evoked from a given context, yet not all questions are equally good for an application.
In theoretical linguistics, 
this also brings up a long-standing gap in understanding how discourse progresses~\cite{warstadt2020just}:
some of such inquisitive ``potential questions'' (as named in \citet{onea2016potential}) are likely more pertinent than others. 
Some of these questions may be answered (by the writer) later in the article and thus become Questions Under Discussion (QUDs)~\cite{roberts2012information}. 
Evidence in psycholinguistics indicate that readers form expectations how a discourse would progress~\cite{kehler-rohde-2017-evaluating}, providing a foundation for the predictability of QUDs~\cite{westera-etal-2020-ted}. \citet{van2003questioning} argues that a question is important if answering it provides high utility. However, there is so far no computational work to 
predict whether or not those questions should be answered or how salient they are.

\begin{figure*}
    \centering
    \includegraphics[width=0.97\textwidth]{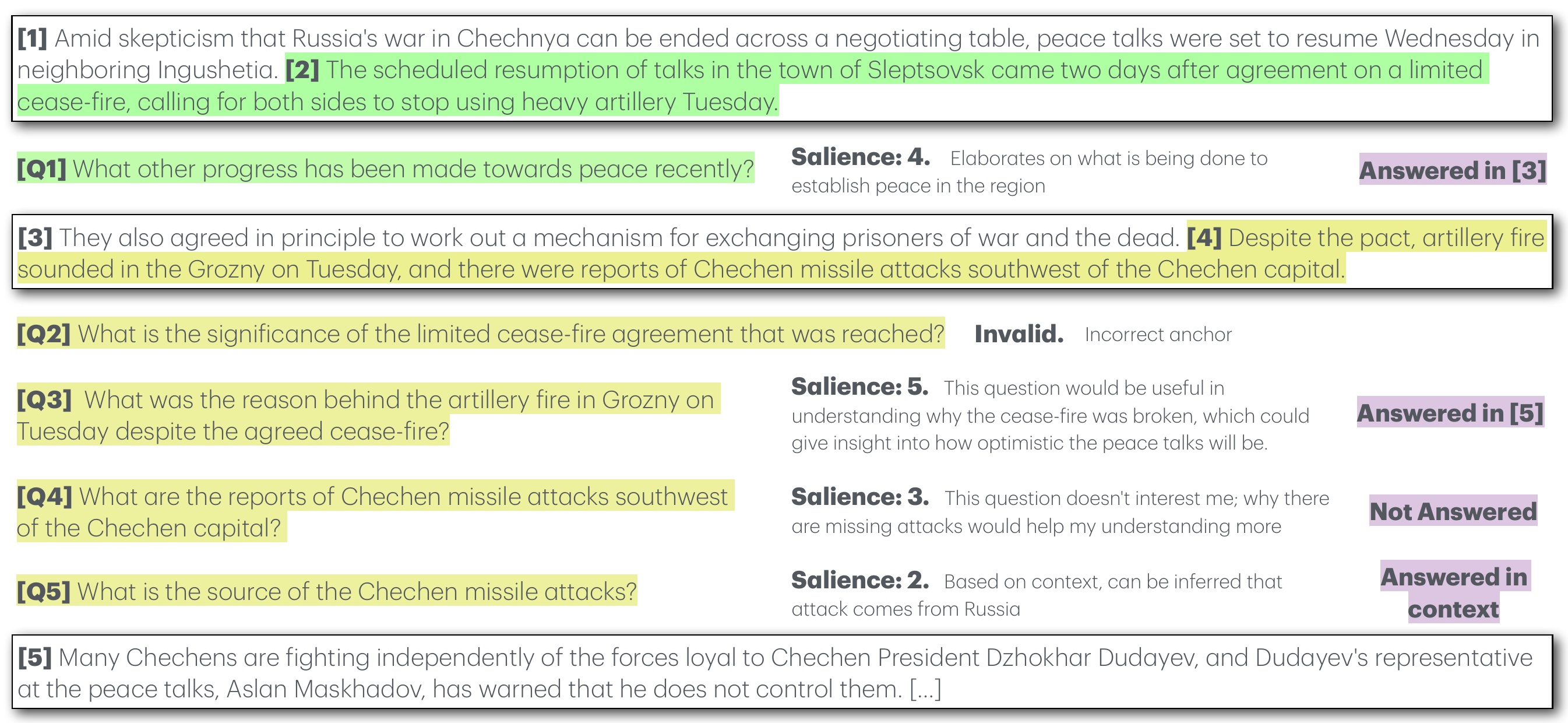}
    \vspace{-0.5em}
    \caption{Examples of inquisitive questions and their annotated salience (with rationales). Each question is evoked by an anchor sentence (shown in the same highlight color). Whether the question is answered is shown on the right. Q1 is taken from human-annotated QUDs in DCQA~\cite{ko-etal-2022-discourse}; Q2-5 are GPT-4 generated questions.} %
    \label{fig:examples}
\end{figure*}

This work (Figure~\ref{fig:examples}) seeks to answer these questions by training a salience prediction model for inquisitive questions, using a new linguist-annotated corpus. In line with \citet{van2003questioning}, a question scores high on salience if answering it would greatly enhance the understanding of the text. First, we collected validity and salience ratings of 1,766 inquisitive questions over English news articles \cite{ko-etal-2022-discourse,huang2023embrace} and TED talks \cite{westera-etal-2020-ted}. A subset of these questions were also annotated in terms of ``answerability'', i.e., how well they were answered by the same article.
Not only do our annotators largely agree with each other on their salience ratings, these ratings also correlate with answerability. Furthermore,
human-generated QUDs from \citet{ko-etal-2022-discourse}, whose answers were guaranteed to be present in the article, also received high salience ratings. These empirical findings support the hypothesis that there is, to some degree, a ``common'' notion of question salience---capturing reader expectations---that connects to answer utility~\cite{van2003questioning}.

We then present \tool, instruction-tuned from open-sourced LLMs that predict salience ratings given an article context. \tool{} is the first application-agnostic question salience model, and outperforms GPT-4 under zero-shot, few-shot,
and Chain-of-Thought prompting \cite{wei2023chainofthought} when evaluated on our dataset.
Encouragingly, even much smaller models such as Flan-T5~\cite{chung2024scaling} achieves significant correlations with human annotated salience. Our experiments show the utility of instruction-tuning on long-context discourse tasks that capture implicit information arising from the cognitive processing of text.

Finally, we take a first step to investigate the value of question salience prediction in a downstream task,
where a short TL;DR of a news article is expanded to a 240-word summary. 
We show that summaries that answer more salient inquisitive questions from the TL;DRs are also ranked higher by human readers.

{\tool} is available at \url{https://github.com/ritikamangla/QSalience}.

\section{Background}\label{sec:background}

\paragraph{Theory: Potential Questions and QUDs}
There are two concepts in linguistics that are relevant to inquisitive questions discussed in this work. First, \citet{onea2016potential} define the semantics of ``potential questions''; informally, a question $Q$ evoked (or licensed) by an utterance $u$ in a given context $c$ such that the answer space of $Q$ is entailed by the common ground of $\{c,u\}$, and that $c$ alone does not license $Q$.
Second, the pragmatics theory of Question Under Discussion (QUD) views discourse as progressively answering questions that were explicitly or implicitly generated in prior context~\cite{van1995discourse,roberts2012information,benz2017questions}. Under the QUD view, a potential question $Q'$ is a QUD if it is answered by an utterance $u'$ where $u'$ is not part of the common ground but is entailed by it (e.g., an upcoming utterance later in an article). In Figure~\ref{fig:examples}, Q1 evoked by sentence 2 is the QUD of sentence 3, and Q3 evoked by sentence 4 is the QUD of sentence 5. Some QUDs evoked earlier in an article can be answered much later \cite{ko-etal-2022-discourse}.

The salience of potential questions and its connection with QUDs, however, is understudied. 
\citet{onea2016potential} listed several hypotheses for the ordering of potential questions, but acknowledged that they are limited and presented no formal or empirical validation. \citet{kehler-rohde-2017-evaluating}'s psycholinguistic experiments showed that people form \emph{expectations} of what QUDs are upcoming using contextual cues, and that such expectations affect their interpretation of linguistic forms. This is compelling evidence for the incremental processing of discourse~\cite{altmann1988interaction,cristea-webber-1997-expectations} and why people ask questions. \citet{westera-etal-2020-ted} later studied how likely a potential question is answered, using the TED-Q corpus that annotates both questions and answers in a (limited) moving window of context. Yet this work focuses on the predictability of QUDs, rather than a reader-centric view of salience as in ours. 
Salience defined in our work is in line with \citet{van2003questioning}'s information-theoretic argument that questions are salient when information utility of the answer is high; 
yet empirical evidence at-scale is yet to be seen.

\paragraph{Applications: Generating Inquisitive Questions}
Prior work developed datasets and models for generating inquisitive questions (defined as open-ended high-level questions targeting discourse understanding) \cite{ko-etal-2020-inquisitive,gao-etal-2022-makes}, which was later used in a range of applications~\cite{laban-etal-2022-discord,wu-etal-2023-elaborative,fok2023qlarify,newman-etal-2023-question,trienes2024infolossqa}. However, this existing work does not explicitly define or model question salience. In QUD parsing, prior work focuses on what makes questions linguistically felicitous QUDs~\cite{riester2018annotation,wu-etal-2023-qudeval}.

A question salience model, however, is often necessary in downstream applications. For instance, in elaborative simplification~\cite{wu-etal-2023-elaborative}, the lack of a salience model means that existing approaches cannot predict which concepts to insert explanations for. Indeed, the over-generation of valid, fine-grained inquisitive questions is undesirable and can easily overwhelm the readers
\cite{trienes2024infolossqa}. In goal-oriented forums, \citet{rao-daume-iii-2018-learning} calculated information utility from the answers to rank clarification questions; however this presupposes an explicit discourse goal to solve a specific task. While domain-specific notions of salience can sometimes be implicitly captured in end-to-end training with a downstream gold-standard (e.g., in summarization planning~\cite{narayan-etal-2023-conditional}), it does not apply to most prior work mentioned above, as they are more open-ended.

\section{Task Definition}\label{sec:salience-schema}

A (human or machine) reader is reading a document, with established context $C_p$ (preceding context) consisting of sentences $\{1,...,S_{k-1}\}$. The reader generates a potential question (Section~\ref{sec:background}; \citet{onea2016potential}) $Q$ evoked at sentence $S_k$ (also called the ``anchor sentence''~\cite{wu-etal-2023-qudeval}).

The salience of $Q$ is the measure of the extent to which it is important for a question $Q$ to be answered, in order to gain a fuller understanding of the situation described, after its invocation at sentence $S_{k}$ \cite{van2003questioning}. Specifically, for all valid questions, we define a Likert scale of 1-5
(full definitions found in Appendix ~\ref{annotation:salience}): 
\vspace{-0.5em}
\begin{itemize}
\itemsep0em 
    \item Score = 1: $Q$ is not related to $C_{p}$
    \item Score = 2: $Q$ is related to $C_{p}$ but seems to be a stretch to ask, and answering it is not useful.
    \item Score = 3: $Q$ is related to $C_{p}$ but whether it is answered does not matter much to the reader.
    \item Score = 4: $Q$ is related to $C_{p}$ and answering it might clarify some newly introduced concepts, or might expand on an idea already introduced.
    \item Score = 5: $Q$ is related to $C_{p}$ and it should definitely be answered as it clarifies a concept introduced in $S_{k}$ or asks for more information about newly introduced humans (or animated) individuals into the discourse.
\end{itemize}
\vspace{-0.5em}
A question is \emph{invalid} if it contains grammatical or factual errors, or is not anchored in $S_{k}$. The last criteria follows linguistic constraints in \citet{wu-etal-2023-qudeval} reflecting that the content of $Q$ is not grounded in $S_{k}$, hence should not be evoked at $S_{k}$.

\paragraph{A note on subjectivity.} The salience values are, to some degree, subjective. However, prior work has shown compelling evidence that certain QUDs are more predictable than others~\cite{westera-etal-2020-ted} and that linguistic cues in the text play a significant role in readers' expectation~\cite{kehler-rohde-2017-evaluating}. Under the assumption that there isn't too much divergence between the authors' intended audience and the background of the actual readers, our work sets out to capture such expectations through a question salience score.

\section{Data Collection}
We first present \data, a corpus of 1,766 inquisitive questions annotated with salience, plus natural language rationales for their judgments. Although question generation has been used widely, application-independent datasets consisting of human-generated inquisitive questions are scarce. Thus, we generate questions with LLMs, both to obtain a sizable amount of data and also to understand inquisitive question generation capabilities of LLMs. We supplement these questions with a smaller number of human-generated questions from prior work, which allows us to perform deeper analysis (Section~\ref{sec:answerability}).

\subsection{Source Texts and Questions}\label{sec:qg}
Table~\ref{tab:avg-std-q} summarizes the number of source articles and questions in \data{}. We draw these from different existing corpora to support different facets of our experimental analysis. They are from:

\textbf{(1)} News texts from DCQA \cite{ko-etal-2022-discourse}.\footnote{\citet{ko-etal-2020-inquisitive} contains human-generated questions, but all the questions are from only the first five sentences of news texts. Thus, annotating them will provide only signals when $C_p$ is very small.} We generate questions from DCQA articles, with gradually increasing size of $C_p$. Additionally, 
the annotated QUDs in DCQA allows us to 
study the salience of QUDs compared to inquisitive questions in general (Section~\ref{sec:answerability}).

\textbf{(2)} TED talks from TED-Q~\cite{westera-etal-2020-ted}. In addition to LLM-generated questions, we also annotate the salience of one of the 6 excerpts with human generated questions in the TED-Q dataset. This provides data for further analysis on question salience vs.\ how answerable they are (Section~\ref{sec:answerability}).

\textbf{(3)} DiverseSumm~\cite{huang2023embrace} contains a newer set of news articles for which 
we annotate salience of LLM-generated questions. For convenience, we denote this subset as \emph{Div.\ Article}. 
These are source articles for  Section~\ref{sec:summ}, our downstream task. To ensure fair evaluation for the task, the articles we selected were all roughly 1,500 words. 
Additionally, we annotate question salience on a set of GPT-4 generated short TL;DRs for these articles. We denote this subset \emph{Div.\ TL;DR}.

\paragraph{Machine Generated Questions}
Given the preceding context $C_{p}$ along with the anchor sentence $S_{k}$, we prompt LLMs to generate 5 questions about a part of the sentence that a reader may be curious about (settings and prompt in Appendix~\ref{sec:qgsettings}).
Multiple LLMs were used to cover a more diverse set of question styles in the dataset. Specifically, 250 questions were generated from Llama-2-7B-chat, 249 from Mistral-7B-instruct, 100 from GPT-3.5-turbo, and 1,106 from GPT-4-turbo.

For full articles, we begin the question generation process from the 4th sentence until the 16th sentence, maintaining a gap of two sentences between consecutive question generation probes, similar to \citet{westera-etal-2020-ted}. 
For the DiverseSumm TL;DRs which are typically 3 sentences long (50 words), we generate questions per sentence.

\paragraph{Human Generated Questions}

We annotate the salience of 61 human generated questions from the above sources, to perform analyses in Section~\ref{sec:answerability}. Among those, 36 of them are derived from 2 articles of {\sc DCQA} and 25 of them are from one article of TED-Q.

\subsection{Salience Annotation}\label{sec:salience-annotation}

\data{} is annotated by three linguistics undergraduate students at our institution who are native English speakers. They have previously been involved in multiple linguistic annotation tasks and have been trained on our specific annotation guideline on 50 questions (25 questions $\times$ 2 articles). The annotation guideline can be found in Appendix \ref{annotation:salience}.\footnote{1,150 questions were annotated by all three annotators; the rest, which is a subset of DiverseSumm articles and TL;DRs, were annotated by two of the three annotators.}
In addition to the labels, annotators also provide natural language rationales which we release with \data{}.
These rationales are used in few-shot Chain-of-Thought prompting~\cite{wei2023chainofthought} in Section~\ref{sec:salience_prediction}.
The annotators were paid at least \$15/hr.

\paragraph{Agreement} The inter-annotator agreement (IAA)
as measured by the Krippendorff's alpha \cite{Krippendorff2011ComputingKA} (ordinal, with the ``invalid'' label set to 0)  is 0.719 for the DCQA articles, 0.632 for TED-Q, 0.751 for Div.Article and 0.649 for Div.TL;DR. These values indicate substantial agreement~\cite{artstein-poesio-2008-survey}, providing evidence to the predictability of reader expectations manifested as inquisitive questions.

\paragraph{Aggregation} For label aggregation, we take the average salience of all annotations, then round it to the closest integer.

\begin{table}[]
\small
\renewcommand{\tabcolsep}{1.4mm}
\centering
\begin{tabular}{@{}lllll@{}}
\toprule
dataset & \#articles & \#questions & \begin{tabular}[c]{@{}l@{}}average\\ length\end{tabular} & \begin{tabular}[c]{@{}l@{}}standard\\ deviation\end{tabular} \\ \midrule
DCQA & 4 & 260 & 11.97 & 4.76 \\
TED-Q & 1 & 100 & 11.07 & 3.9 \\
Div. Article & 27 & 957 & 14.8 & 4.44 \\
Div. TL;DR & 34 & 449 & 16.32 & 3.78 \\ \midrule
All & 66 & 1766 & 13.99 & 4.57 \\ \bottomrule
\end{tabular}
\caption{Count of articles and questions, average length and standard deviation of human and machine-generated questions per dataset.}
\label{tab:avg-std-q}
\vspace{-0.5em}
\end{table}

\begin{table}[]
\centering
\small
\begin{tabular}{@{}lrrrrrr@{}}
\toprule
dataset      & Invalid    & 1    & 2    & 3    & 4    & 5    \\ \midrule
DCQA  & 13.4 & 2.3 & 19.2 & 36.9 & 20.0 & 8.0 \\
TED-Q        & 14.0 & 0 & 6.0 & 29.0 & 35.0 & 16.0 \\
Div. Article & 9.9 & 0.8 & 17.2 & 22.6 & 31.9 & 17.6 \\
Div. TL;DR & 3.6 & 0.2    & 4.7 & 12.0 & 47.0 & 32.5 \\ \midrule
All & 9.1 & 0.8 & 13.7 & 22.4 & 34.1 & 19.9\\\bottomrule
\end{tabular}
\caption{Validity and salience distribution (in \%) of human-annotated labels for the questions in \data{}.}
\label{tab:distribution}
\end{table}

\paragraph{Analysis} Examples of the annotated data are shown in Figure~\ref{fig:examples} and Appendix Table~\ref{tab:salience-example}. Table \ref{tab:distribution} provides the label distribution for \data{}. Notably, more than 90.8\% 
of the questions generated from LLMs are valid, making them promising tools for inquisitive question generation. Our qualitative analysis of annotator rationales for invalid questions show that many of them does not have the right anchor sentence (i.e., $Q$ not anchored in $S_k$); this was also found in \citet{wu-etal-2023-qudeval}. A few invalid questions also contain non-factual presuppositions. 
Among valid questions, those with the lowest score of 1 (question was irrelevant to $C_p$) is rare. However, the salience of the questions varies, indicating the potential usefulness of a salience predictor for LLM-generated questions in downstream tasks.
We further analyze salience scores stratified by the LLMs that generate the questions in Appendix~\ref{sec:modelsalienceanalysis}.

In Appendix~\ref{app:qtype}, we show that question types that are more likely to associate with high salience ratings are \emph{Consequence, Example}, and \emph{Procedural}. \emph{Disjunctive, Concept} and \emph{Judgmental}.

\begin{table}[t]
\centering
\small
\begin{tabular}{@{}lllll@{}}
\toprule
dataset     & 0    & 1    & 2    & 3    \\ \midrule
DCQA & 0.28 & 0.35 & 0.16 & 0.21 \\
TED-Q       & 0.07 & 0.22 & 0.23 & 0.48 \\ \bottomrule
\end{tabular}
\caption{Distribution (in \%) of human-annotated answerability labels for 311 questions stratified by data source.}
\label{tab:distribution-ans}
\end{table}

\section{Salience vs.\ Answerability}\label{sec:answerability}

A valid potential question evoked at $S_k$ can be deemed a QUD anchored at $S_k$ if the subsequent discourse answers it. In order to study the relationship between potential questions and QUDs, we 
annotate a subset of the questions in \data{} in terms of their \emph{answerability},
i.e., the extent to which $Q$ is answered in the subsequent context $C_{s}$.\footnote{Although \citet{westera-etal-2020-ted} annotated this (they call this QUD predictability), we did not use their labels because they annotated whether a question was answered within a window of 4 sentences after $S_k$ rather than the full texts.}
We annotate answerability
on a Likert scale: fully answered (3), partially answered (2), unanswered by $C_{s}$ (1), and already answered in $C_p+S_k$ (0).
The same annotators as in Section~\ref{sec:salience-annotation}  annotated 
225 and 86 valid questions from the DCQA and TED-Q subsets,
respectively. Krippendorff's alpha (ordinal) is 0.799, indicating substantial agreement. Table \ref{tab:distribution-ans} shows the distribution of answerability scores. Appendix~\ref{sec:modelanswerabilityanalysis} provides more analysis on answerability scores per question generation model.

\begin{table}[]
\centering
\small
\begin{tabular}{l|lll}
\toprule
Random & \multicolumn{3}{c}{Human Ratings}\\
Questions & All & DCQA & TED-Q \\ \midrule
-0.02* & 0.65 & 0.59 & 0.74 \\ \bottomrule
\end{tabular}
\caption{Spearman rank correlation between salience and answerability annotated by humans and a random baseline. The correlation values that are not statistically significant ($p<0.05$) are marked with a *.}
\label{tab:notions-correlation}
\vspace{-0.5em}
\end{table}

\paragraph{Do salience and answerability correlate?}
Table \ref{tab:notions-correlation} presents the Spearman rank correlation coefficients between annotated salience and answerability. 
As comparison, we also compute a random correlation baseline between salience and the answerability of a random question, averaged across 10 trials. 
The annotated salience and answerability values have a significant Spearman's $\rho$ of 0.65 (compared to the random baseline of $-0.02$).  \textbf{This is evidence suggesting that salient potential questions are also likely to be answered later in the discourse,} even though the writers do not observe reader questions. This suggests that reader and writer expectations align to a certain degree.

\paragraph{Are QUDs salient potential questions?}
Table \ref{tab:dcqa-human-sal-dist} presents the salience distribution of 36 DCQA questions that are annotated QUDs.
Similar to the previous analysis, we also take a random subset of 36 potential questions, averaging their scores over 10 trials and present their salience distribution. We see that QUDs, which are potential questions answered in later context, are overall much more salient compared to a random set of potential questions sampled from \data.

\begin{table}[]
\small
\centering
\begin{tabular}{@{}llllll@{}}
\toprule
 & 1 & 2 & 3 & 4 & 5 \\ \midrule
Random Questions & 0.01 & 0.20 & 0.39 & 0.28 & 0.12 \\
Annotated QUDs & 0 & 0.11 & 0.25 & 0.47 & 0.17 \\ \bottomrule
\end{tabular}
\caption{Salience distribution for 36 human annotated QUDs from DCQA, compared to a random set of inquisitive questions of the same size.}
\label{tab:dcqa-human-sal-dist}
\end{table}

\section{Salience Prediction}\label{sec:salience_prediction}

\begin{table*}[!ht]
\small
\centering
\begin{tabular}{lcccccc}
    \toprule
    \textbf{Model}  & \textbf{MAE ↓} & \textbf{Spearman ↑} & \textbf{Macro F1 ↑} & \textbf{krippendorff's $\alpha$ ↑} \\
    \midrule
    GPT4 zero-shot (vanilla)&  1.314 & 0.229 & 0.193 & -0.141\\
    GPT4 few-shot (vanilla)&  \cellcolor{blue!25}0.910 & \cellcolor{blue!25}0.417 & \cellcolor{blue!25}0.316 & \cellcolor{blue!25}0.358 \\
    GPT4 few-shot (kNN)& 1.063 & 0.359 & 0.245 & 0.215 \\ 
    GPT4 CoT zero-shot &  1.144 & 0.366 & 0.197 & 0.058\\
    GPT4 CoT few-shot &  1.034 & 0.327 & 0.292 & 0.165\\
    \midrule
    \tool{} (Mistral-7B-instruct) &\cellcolor{blue!25}\textbf{0.579} & \cellcolor{blue!25}\textbf{0.623} & \cellcolor{blue!25}\textbf{0.417} & \cellcolor{blue!25}\textbf{0.615} \\
    Llama-2-7B-chat & \textbf{0.626} & \textbf{0.566} & \textbf{0.413} & \textbf{0.557} \\
    Flan-t5-base  & 0.706 &  0.542 & 0.370 & 0.526 \\
    TinyLlama-1.1B-chat & 0.664 &  0.522 & 0.402 & 0.496 \\

    \bottomrule
\end{tabular}
\caption{Model performance on the salience prediction task, for GPT-4 zero/few-shot baselines (top) and instruction-tuned LLMs (bottom). \textbf{Bold}: top-2 performance; blue shades: best performance for baselines and for fine-tuned models.}
\label{tab:model-performance}
\vspace{-0.5em}
\end{table*}

We experiment with a range of models for the prediction of question salience, given valid questions. Our finding is that salience prediction is 
a discourse task that recovers implicit information not readily grasped by LLMs, while our best instruction-tuned model, \tool{}, can achieve moderate agreement with humans.
We further present question validity classifiers in  Appendix \ref{sec:binary-cls}, which can be used with \tool{} in a pipeline fashion.

\subsection{Models}

\paragraph{Instruction Tuning}
We instruction fine-tune several open-source LLMs with QLoRA \cite{dettmers2024qlora}: \textbf{Mistral} \cite{jiang2023mistral} ({\tt Mistral-7B-Instruct}), \textbf{Llama 2} \cite{touvron2023llama} ({\tt Llama-2-7b-chat}), \textbf{TinyLlama} \cite{zhang2024tinyllama} ({\tt TinyLlama-1.1B-chat}), and \textbf{Flan-T5} \cite{chung2024scaling} ({\tt flan-t5-base}). AdamW~\cite{loshchilov2018decoupled} is used for optimization. Hyperparameters can be found under Table \ref{tab:training_params} in the Apendix.

The training data is formulated as (\emph{input, output}) pairs where \emph{input} consists of $C_p, S_k, Q$, and \emph{output} is the salience score.\footnote{We also tried fine-tuning a classification head; however performance is inferior.} Appendix~\ref{app:fine-tune-setting} shows the instructions for these models. For Flan-T5, since the context span is 512 tokens, we also experiment without using instructions, and truncate the context in the reverse sentence order from $C_p$ and $S_k$ until the context length is filled.

\paragraph{LLM Zero-/Few-shot Baselines}
We perform extensive experiments with various zero-shot and in-context learning scenarios with 
\textbf{GPT-4}\texttt{-turbo}. We show prompts in Appendix \ref{app:gpt-salience-prompts}.

(1) \textbf{Zero-shot (vanilla)}, where the model is given an instruction similar to that of the annotators.

(2) \textbf{Few-shot (vanilla)}, where 15 in-context learning examples (3 per label) of $((C_p, C_k, Q), scr)$ are given, where $scr$ denotes the salience score. 
We utilize LLMs' recency bias~\cite{liu2023lost} to nudge its prediction to better align with our label distribution. Thus we altered the order of in-context demonstrations such that the examples at the end have labels more frequent within our train set.

(3) \textbf{Few-shot (kNN).}
Performance of LLMs is often sensitive to the selection of the in-context examples \cite{rubin-etal-2022-learning}. Hence we use a kNN-based approach \cite{liu2021makes} to find the closest in-context examples to the current test instance. We encode $C_{p}$ and $S_{k}$ separately with RoBERTa-large~\cite{liu2019roberta} and take the average of the CLS tokens of each. We use Euclidean distance and retrieve one closest example for each salience label. These examples are put in-context following a similar ordering as the few-shot (vanilla) setting.

(4) \textbf{Chain-of-Thought (CoT).} 
We experimented with Chain-Of-Thought prompting \cite{wei2023chainofthought}. %
For \textbf{few-shot CoT}, we use 5 in-context examples, with the reasoning taken from the natural language rationales that the annotators gave during salience annotation.

\subsection{Evaluation}

\paragraph{Data} We create a test set of 235 valid questions. The rest of the dataset is split into training (1,228 valid questions) and validation (143 valid questions). The data splits are stratified by articles. We upsample the training data to balance the label distribution. Our final training set consists of 2,355 examples, where each label has its 471 examples. We do not upsample validation or test sets.

\paragraph{Evaluation Metrics}
We measure the performance of salience prediction using four metrics: \textbf{(1)} Mean Absolute Error (MAE) between the predicted salience scores and the aggregated human scores; \textbf{(2)} Spearman's $\rho$ between the two; \textbf{(3)} macro-averaged F1. These are standard metrics for ordinal classification or regression. In addition, we report \textbf{(4)} Krippendorff's $\alpha$ that measures agreement, also used in Section~\ref{sec:salience-annotation} between annotators.

\paragraph{Results} 
Table~\ref{tab:model-performance} shows that the fine-tuned models clearly outperform zero- or few-shot LLMs, even with stronger prompting techniques such as kNN-based in-context example retrieval and Chain-of-Thought.
On the contrary, among the fine-tuned smaller models, the best performing Mistral-based model achieves moderate agreement with human annotation with a substantial correlation. Even \texttt{flan-T5-base} with only 250M parameters and a small context window can be fine-tuned for this task to achieve competitive performance. These conclusions indicate that question salience is difficult to elicit from LLM prompting or in-context learning, and that explicit training can successfully capture this notion.

Appendix Figure \ref{fig:confusion_matrix} shows the confusion matrix for zero-shot GPT-4, the best-performing GPT-4 setting (few-shot vanilla), and our fine-tuned models.\footnote{Note that the label 1 is extremely rare (Table~\ref{tab:distribution}) and is not present in the test set.} 
Compared to fine-tuned models, GPT-4 tends to give a high score for the question by predicting many 4s and 5s, indicating its inability to distinguish good vs.\ bad questions irrespective of the Likert scale. By comparison, predictions from fine-tuned models tend to confuse primarily  labels closer to each other. This also shows our fine tuned models 
understand the tasks better than in-context learning with GPT-4.

\begin{figure*}
    \centering
    \includegraphics[width=0.97\textwidth]{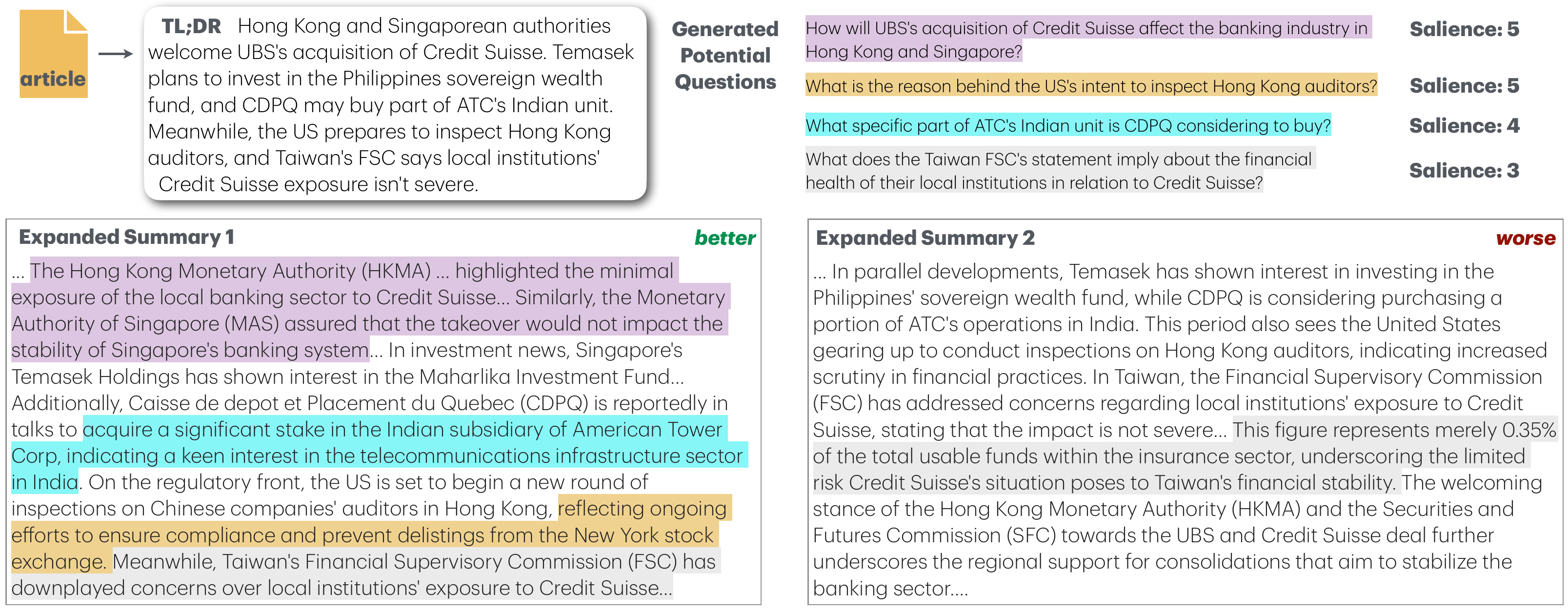}
    \vspace{-0.5em}
    \caption{This example illustrates the expanded summarization task and how question salience is used in evaluation. The summaries are generated by expanding on a short TL;DR of an article. We show the inquisitive questions generated from the TL;DR and their salience scores. A better summary (left) answers more salient questions than the worse summary (right), where only one medium-salient question is answered.}
    \label{fig:summ_example}
\end{figure*}

\section{Use Case: Do better expanded summaries answer more salient questions?}\label{sec:summ}
We demonstrate the usefulness of question salience prediction in a downstream task: summary expansion. Given a document $D$ and a short TL;DR $S_s$, the summary expansion task aims to generate a longer summary $S_l$ that captures a fuller picture of the article, shown in Figure~\ref{fig:summ_example}. This task captures the situation where after reading the TL;DR, a curious reader often wants to know more in order to decide if they want to read the entire article. A similar task is deployed in 
Semantic Scholar, though in a very different domain than ours.

Given findings in Section~\ref{sec:salience-annotation}, our hypothesis is that reader expectation aligns with QUDs in the expanded summary; namely, a higher-quality summary answers more salient questions.

\subsection{Data}
\paragraph{Articles and TL;DRs} We sample 34 articles from DiverseSumm as source articles; these articles are roughly 1,500 words long. Given the strong performance of GPT-* in summarization \cite{goyal2022news,zhang2023benchmarking}, we prompt GPT-4\texttt{-turbo} to produce a 50-word summary TL;DR of the article.
All prompts needed in this section are shown in Appendix \ref{app:summary-exp-prompts}.

\paragraph{Expanded Summaries} For each TL;DR, we generate three expanded summaries while controlling their lengths to be between 230--250 words:

\textbf{(1) GPT-4:} given $(D,S_s)$, we prompt GPT-4\texttt{-turbo} for $S_l$. 

\textbf{(2) Flan-T5:} we use \texttt{flan-t5-large} to produce an elaboration in a similar manner as above.\footnote{We prompt the model with the instruction, TL;DR and the article in that order. Owing to the short context size of \texttt{flan-t5-large}, the entire article is not used for the task.} Since this model can produce summaries with obvious errors such as repeating sentences and violating the length control, we refine the Flan-T5 outputs to fix these obvious errors, while preserving the summary as much as possible, using GPT-4\texttt{-turbo}.

\textbf{(3) GPT-4-Corrupted:} We synthetically generate ``bad'' summaries to serve as a baseline that is missing the most prevalent topics. 
First, we prompt GPT-4\texttt{-turbo} with $(D, S_s)$ and ask it to identify important topics from $S_s$.
Using these topics and $D$, we then prompt the model to generate a long summary within our expected word count that does not include these relevant topics.
Since this response can sometimes be incoherent and disobey the length control, so we again use GPT-4\texttt{-turbo} to refine the output with the same refinement prompt as Flan-T5.

\paragraph{Question Salience} Each sentence in every TL;DR is associated
with 5 LLM generated inquisitive questions (Section~\ref{sec:qg}), annotated with salience (Section~\ref{sec:salience-annotation}). Additionally, we run fine-tuned models from Section~\ref{sec:salience_prediction} for all valid questions in this set to obtain predicted salience values.

\subsection{Evaluation}
\paragraph{Human Ranking of Summaries}
We recruited 5 experienced annotators to rank the expanded summaries, focusing on their content rather than style.\footnote{Prior work showed that GPT3-generated summaries are faithful \cite{zhang2023benchmarking}.} 
Each of the three summaries is given a score from 1 (lowest) to 3 (highest) based on their rankings. 
Our annotation interface can be seen in Figure \ref{fig:annotation_interface} in the Appendix. Examples are shown in Figure~\ref{fig:summ_example} and Table~\ref{tab:elab:full:example}.

The ranking results show an expected order of quality: GPT-4 (2.91 average ranking), Flan-T5 (1.73), GPT-4-Corrupted (1.35). This is the oracle ordering that we aim to reproduce by utilizing the salience of QUDs in the expanded summary.

\begin{table}[t]
\small
\centering

\begin{tabular}{@{}llll@{}}
\toprule
 & GPT-4 & Flan-T5 & GPT-4-Crpt \\ 
 \midrule
Human Salience & 26.9 & 18.3 & 17.8 \\
Mistral-7B-instruct & 29.0 & 20.2 & 19.2 \\
Llama-2-7B-chat & 29.4 & 21.0 & 19.8 \\
Flan-T5-base & 30.8 & 21.8 & 20.8 \\ 
TinyLlama-1.1B-chat & 29.0 & 20.6 & 18.8 \\ \bottomrule
\end{tabular}
\caption{SummScr of human and model salience for 5 TL;DR expansion instances}
\label{tab:elab-salience-scores}
\end{table}

\paragraph{Measuring QUD Salience}
To score a summary $S_l$, we measure the salience $scr$ of the questions evoked in each TL;DR $S_s$ that become QUDs (i.e., answered in $S_l$).
First, we filter all questions that are not answered by the article $D$ itself.%
Next, with the remaining questions and $S_l$, we prompt GPT-4-\texttt{turbo} to return all the questions that $S_l$ answers.
Prompts in this section are shown in Appendix \ref{app:summary-eval-prompts}.

The salience scores $scr(q)$ of these QUDs are then aggregated into
{\tt SummScr} $=\frac{1}{n}\sum_{i=1}^{n}{\sum_{q\in Q_i}{scr(q)}}$
where $Q_i$ denotes all answered questions in the generated expanded summary $i$.

\paragraph{Results}
The SummScr (calculated from human salience) for all 34 articles is GPT-4 (21.93), Flan-T5 (16.25) and GPT-4-Corrupted (8.81). The Kendall's $\tau$ rank correlation between the majority summary ranking and SummScr (human) for the full set is 0.529, a moderately high correlation. 

Table \ref{tab:elab-salience-scores} shows the SummScr on the test set portion of DiverseSumm. All SummScr values derived from \emph{predicted} salience values, using fine-tuned systems, reproduce the same system-level ranking as humans:
GPT-4, Flan-T5, then GPT-4-Corrupted. We consider this evidence that better expansions answer more salient questions.

\section{Conclusion}

In this paper, we explored predicting salience for inquisitive questions. Our work connects two ideas: a theoretical idea of which questions are useful for understanding and likely to be answered later in a text, and an empirical notion of what questions are useful. We showed that predicting salience is possible with fine-tuned models, and these approaches outperform GPT-4. Furthermore, we showed in a pilot use case that notions of summary quality align with how many salient questions were answered.

\section*{Limitations}

While this work takes the first step at empirically connecting prior discourse literature and developing a salience model for inquisitive questions, we have not engaged in the formal semantics of potential questions as in \citet{onea2016potential}. An additional limitation is that we have not explicitly measured information utility (in information-theoretic terms) given the open-ended nature of the questions, although our notion of salience is consistent with \citet{van2003questioning}.

This work has considered only English text, sourcing articles from existing datasets that provided groundwork for various analyses in this paper, both theoretical ones and empirical experiments. Thus even though our notion of question salience is application-agnostic, we believe an exciting future direction is to explore question salience in other domains and languages.

Finally, when considering the notions of salience for our texts, we assume that the reader backgrounds are not too divergent from what the writer has intended. A large discrepancy between the two could lead to readers having very different salient questions; e.g., when the reading level of the reader is much lower than that of the intended audience~\cite{wu-etal-2023-elaborative}. Thus our tool and dataset should not be used when 
reader backgrounds are too different from the writer expectations or among themselves.

\section*{Acknowledgments}
Special thanks to Kathryn Kazanas, Keziah Reina, Karim Villaescusa F, Akhila Gunturu, Andrea Conde, Jada Li and Melanie Quintero for providing data annotation for this project. This research has been supported by NSF Grants IIS 2145479, IIS 2145280, AF 1901292, CNS 2148141, Tripods CCF 1934932, IFML CCF 2019844, a grant from
Open Philanthropy, and research gifts by Western Digital, Amazon, WNCG IAP, UT Austin Machine Learning Lab (MLL), Cisco, and the Stanly P. Finch Centennial Professorship in Engineering.
\bibliography{custom}

\clearpage
\appendix

\section{Annotation Guidelines for Question Salience}\label{annotation:salience}

\textbf{Motivation} As one reads an article, it is natural to ask curiosity-driven questions to enhance one’s understanding of the article. Amongst different potential questions that one might ask while reading the article, to what extent is it important for it to be answered later in the article? Can we perhaps rank these questions? We develop an evaluation schema to do just that!

\textbf{Task} Given the prior context, anchor sentence, and a list of potential questions, score the questions on the basis of the following schema.

\textbf{Score=0}: These are questions which satisfy atleast one of the following criterion:
\begin{enumerate}
\itemsep0em
    \item Question has grammar errors
    \item Question is not anchored in the given anchor sentence
    \item Question contains multiple sub-questions
    \item Question misinterprets the context
\end{enumerate}

\textbf{Score=1}: The question is not very related to the topic (basically to weed out any odd questions)

\textbf{Score=2}: The question is related to the concepts introduced in the prior context and the anchor sentence but asking the question seems like a stretch. Answering the question doesn't seem useful in making the article feel complete. Typically questions that also seem to be completely answered by the prior context and the anchor sentence are given this score.

\textbf{Score=3}: The question is related to the prior context and anchor sentence but answering it doesn't matter to me. Answering it may provide additional information which may/may not enhance my understanding of the article.

\textbf{Score=4}: Answering the question is somewhat useful because, for example, it might clarify some newly introduced concepts, or might expand on an idea already introduced. It is useful to answer the question because it might influence the narrative. There is a degree of uncertainty here as compared to when you would score a question 5.

\textbf{Score=5}: This question should definitely be answered in the subsequent context. Some reasons why the question should definitely be answered:
\begin{enumerate}
\itemsep0em
    \item It clarifies a concept introduced in the anchor sentence
    \item It asks about surprising or mysterious events/objects
    \item It asks for more information about newly introduced humans (or animated) individuals into the discourse
    \item Answering this question is essential to understanding the narrative.
\end{enumerate}

Do keep in mind that one shouldn't make an inference about other people. For instance, if the question is about defining or explaining a concept, and you don’t need that explanation, don’t say that answering the question may still be helpful just because you think some other people will find the answer useful.

\section{Validity Classification}

Per Table~\ref{tab:distribution},
invalid questions accounted for 9.1\% of the annotated data. Thus we also experiment with question validity classification, which can be used in a pipeline to first find invalid questions to exclude, before scoring their salience.

\paragraph{LLM Zero-/Few-shot Baselines} Since many invalid questions are caused by anchor issues (Section~\ref{sec:salience-annotation}), 
we use the anchor relevance prompt in QUDEval \cite{wu-etal-2023-qudeval} 
for few-shot prompting using GPT-4 and Mistral-7B-instruct to classify question validity.\footnote{We merge ``fully grounded'' and ``partially grounded'' in their label set as a single valid label}. 

\paragraph{Fine-Tuning} We also fine tune {\tt flan-t5-base} and {\tt TinyLlama-1.1B-chat} on this task. Prompt \ref{prompt:binary:model:ft} list the instruction for TinyLlama-chat.\footnote{Due to the small context window of Flan-T5, we do not use instructions.} 
AdamW \cite{loshchilov2018decoupled} was used as optimizer with a learning rate of 3e-4, trained for 2 epochs. We perform downsampling to balance the data distribution.

\paragraph{Results} Table \ref{tab:binary-cls-model-perf} shows that both prompting and fine-tuned models perform decently well on question validity classification. The fine-tuned models are on-par with prompting LLMs.

\label{sec:binary-cls}
\begin{table}[!h]
\small
\centering
\begin{tabular}{lcccccc}
    \toprule
    \textbf{Model}  & \textbf{Macro F1 ↑}  \\
    \midrule
    
    GPT-4 few-shot & 0.689  \\
    Mistral-7B-instruct few-shot& 0.538 \\
    Flan-t5-base fine tuned& 0.662 \\
    TinyLlama-1.1B-chat fine tuned & \cellcolor{blue!25}0.693 \\
    \bottomrule
\end{tabular}
\caption{Question Validity Performance}
\label{tab:binary-cls-model-perf}
\end{table}

\begin{prompt}[title={\thetcbcounter{} Instruction for in fine-tuned models for question validity classification.},label=prompt:binary:model:ft]
\promptsubsection{System}Below is an instruction that describes a task, paired with an input that provides further context. Write a response that appropriately completes the request.

\promptsubsection{Instruction}Is the question well-grounded in the anchor sentence? Please evaluate using the following scale:\\
1: The question is fully grounded in the anchor sentence. Or some parts of the question are grounded in the anchor sentence.\\
0: The question is not grounded at all in the anchor sentence.\\
Based on the question and the anchor, please choose one of the above options. If the question refers to the same entity as the anchor, we consider the question to be grounded. 

\promptsubsection{Input}\\
question: \highlight{question} \\
anchor sentence: \highlight{anchor sentence $S_k$} \\

\promptsubsection{Response}\newline \highlight{score} 
\end{prompt}

\section{LLM Question Generation and Additional Analyses}

\subsection{Settings}\label{sec:qgsettings}
We show the prompt for LLM inquisitive question generation in Prompt \ref{prompt:question:generation}.
A temperature of 0 is used so that the questions generated are grounded within the context provided. Greedy decoding is used due to its computational efficiency and deterministic behaviour. We also use a frequency penalty of 0.5 to make the model more conservative in generating repetitive tokens.

\begin{prompt}[title={\thetcbcounter{} Prompt for Question Generation},label=prompt:question:generation]

\promptsubsection{Context} \\
\highlight{article context $C_p$}

After reading the sentence \highlight{anchor sentence $S_k$}, ask 5 questions about a part of this sentence that you are curious about which you don't have an answer for.
\end{prompt}

\begin{table}[t]
\small
\centering
\begin{tabular}{@{}lll@{}}
\toprule
 & \#questions & avg salience \\ \midrule
human & 61 & 3.49 \\
mistral-7B-instruct & 249 & 2.26 \\
llama-2-7B-chat & 250 & 2.6 \\
GPT-3.5-turbo & 100 & 3.18 \\
GPT-4-turbo & 1106 & 3.75 \\ \bottomrule
\end{tabular}
\caption{Count and average human salience of questions stratified by the question generation model used}
\label{tab:model-sal-dist}
\vspace{2em}

\begin{tabular}{@{}lll@{}}
\toprule
 & \#questions & avg answerability \\ \midrule
human & 58 & 2.25 \\
mistral-7B-instruct & 73 & 1.32 \\
llama-2-7B-chat & 88 & 1.15 \\
GPT-3.5-turbo & 92 & 1.52 \\ \bottomrule
\end{tabular}
\caption{Distribution and average human answerability of questions stratified by question generation model}
\label{tab:model-ans-dist}
\end{table}

\subsection{Salience Analysis Per Model}\label{sec:modelsalienceanalysis}
Table \ref{tab:model-sal-dist} shows the average salience scores of questions produced by each model. Qualitative analysis of annotator rationales reveals that
both Mistral-7B-instruct and Llama-2-7B-chat struggled to generate questions anchored in $S_{k}$, resulting in many invalid questions; Mistral-7B-instruct was worse than Llama-2-7B-chat. On the other hand, GPT-* models produced more interesting, curiosity driven questions. 

\subsection{Answerability Analysis Per Model}\label{sec:modelanswerabilityanalysis}

Table \ref{tab:model-ans-dist} shows the count and average answerability of valid potential questions stratified by the question generating model. Qualitative analyses of annotator rationales reveals that Mistral-7B-instruct and Llama-2-7B-chat often produced questions which were already answered in $C_{p}$ + $S_{k}$ or were about specific parts of $S_{k}$, not relevant to the article as a whole; llama was generally worse than mistral. While GPT-3.5-turbo is good at generating salient questions, they sometimes becomes too diverse to be actually answered by the article. 
Human questions are QUDs and thus are mostly answered.

\section{Prompts for GPT-4 salience prediction}\label{app:gpt-salience-prompts}

GPT-4 zero-shot salience prompts are shown in Prompt \ref{gpt:zero:shot:salience}. For few-shot, 3 examples are taken for each of the 5 labels using the same format. Similarly, zero-shot Chain-of-Thought prompt is shown in Prompt \ref{gpt:few:shot:salience}. Few-shot Chain-of-Thought uses the same format with 3 examples for each of the 5 labels.

\begin{prompt}[title={\thetcbcounter{} Zero-shot (vanilla) prompt for salience prediction.},label=gpt:zero:shot:salience]

\promptsubsection{article}
\highlight{article context $C_p$ + anchor sentence $S_k$}

\promptsubsection{question} \highlight{question}

\promptsubsection{system}Imagine you are a curious reader who is reading the article. You come across a question and you need to determine if it should be answered in the following article or not. You have to give a score for this input. Score = 1 means the question is completely unrelated to the topic of the article. Score = 2 means the question is related to the article but it has already mostly been answered by the article. Score = 3 means the question is related to the article but answering it is not useful as it might expand of an idea that is not very important or central to the context of the article. Score = 4 means the question is related to the article and answering it is somewhat useful in enhancing the understanding of the article. Score = 5 means the question is related to the article and should definitely be answered because it expands on some ideas which are central to the article. Note that the score is given according to the information utility of its answer. If a question is related to the article but doesn't need to be answered or is not central to the article, do NOT give it a high score of 4 or 5, instead give a score of 3 if the question is unanswered by the article and 2 if it has already been answered by the article. To differentiate between a score of 4 vs 5, think of how the article would look like if you don't answer the question - if the article would not feel complete without the answer to the question, give a score of 5, else a 4. A score of 4 is usually given if answering the question will be useful but there might be other questions that are more important to answer as compared to this. A score of 5 is only given to the best and most important questions that MUST be answered so use it carefully and sparingly. Do not be biased towards giving a high score and follow the above instructions carefully. The score should strictly be an integer from 1 to 5.\\
\promptsubsection{score}
\end{prompt}

\begin{prompt}[title={\thetcbcounter{} Zero-shot CoT prompt for salience prediction.},label=gpt:few:shot:salience]

\promptsubsection{article}\highlight{article context $C_p$ + anchor sentence $S_k$}

\promptsubsection{question}\highlight{question}

\promptsubsection{system}Imagine you are a curious reader who is reading the article. You come across a question and you need to determine if it should be answered in the following article or not. You have to give a reason and a score for this input. Score = 1 means the question is completely unrelated to the topic of the article or misinterprets the context of the article. Score = 2 means the question is related to the article but it has already mostly been answered by the article. Score = 3 means the question is related to the article but answering it is not useful as it might expand of an idea that is not very important or central to the context of the article. Score = 4 means the question is related to the article and answering it is somewhat useful in enhancing the understanding of the article. Score = 5 means the question is related to the article and should definitely be answered because it expands on some ideas which are central to the article. Note that the score is given according to the information utility of its answer. If a question is related to the article but doesn't need to be answered or is not central to the article, do NOT give it a high score of 4 or 5, instead give a score of 3 if the question is unanswered by the article and 2 if it has already been answered by the article. To differentiate between a score of 4 vs 5, think of how the article would look like if you don't answer the question - if the article would not feel complete without the answer to the question, give a score of 5, else a 4. A score of 4 is usually given if answering the question will be useful but there might be other questions that are more important to answer as compared to this. A score of 5 is only given to the best and most important questions that MUST be answered so use it carefully and sparingly. Do not be biased towards giving a high score and follow the above instructions carefully. First provide a reasoning for your response and then the score. Now let's think step by step.

\promptsubsection{reason}
\end{prompt}

\section{Setups for instruction fine-tuning}\label{app:fine-tune-setting}
\begin{table}[!h]
\small
\centering
\begin{tabular}{lccccc}
    \toprule
    \textbf{Model} & \textbf{Seq Len} & \textbf{Learn. Rate} & \textbf{Epoch} \\
    \midrule
    Mistral-7B-instruct & 4096 & 0.0003 & 3 \\
    Llama-2-7B-chat &  4096 & 0.0001 & 5 \\
    Flan-t5-base &  512 & 0.0003 & 3 \\
    TinyLlama-1.1B-chat & 4096 & 0.0003 & 4 \\
    \bottomrule
\end{tabular}
\caption{Parameters for fine-tuned Models in salience scoring}
\label{tab:training_params}
\end{table}

\begin{prompt}[title={\thetcbcounter{} Instruction for fine-tuned models for salience prediction.},label=prompt:model:ft]
    \promptsubsection{System}Below is an instruction that describes a task, paired with an input that provides further context. Write a response that appropriately completes the request.
    
    \promptsubsection{Instruction}Give a score from 1 to 5 for how important it is for the question to be answered later in the article.\\
Score = 1 means the question is completely unrelated to the topic of the article.\\
Score = 2 means the question is related to the article but answering it is not useful in making the article feel complete. \\
Score = 3 means the question is related to the article but answering it might not enhance the understanding of the article. \\ 
Score = 4 means the question is related to the article and answering it is somewhat useful in enhancing the understanding of the article. \\
Score = 5 means the question is related to the article and should definitely be answered because it might provide explanation for some new concepts. 

    \promptsubsection{Input}\\
    article: \highlight{article context $C_p$ + anchor sentence $S_k$} \\
    question:\highlight{question}
    
    \promptsubsection{Response:} \newline \highlight{score}
\end{prompt}

\section{Prompts for Expanded Summary Generation}\label{app:summary-exp-prompts}

\begin{prompt}[title={\thetcbcounter{} Prompt for generating a short TL;DR.},label=prompt:short:summ]
    \promptsubsection{Context}\highlight{article}

    Generate a short 50-word summary for the above article. Remember, do not exceed 50 words.

    \promptsubsection{Summary}
\end{prompt}

\begin{figure*}[ht]
    \centering

    \begin{minipage}{\textwidth}
        \centering
        \includegraphics[width=0.48\textwidth]{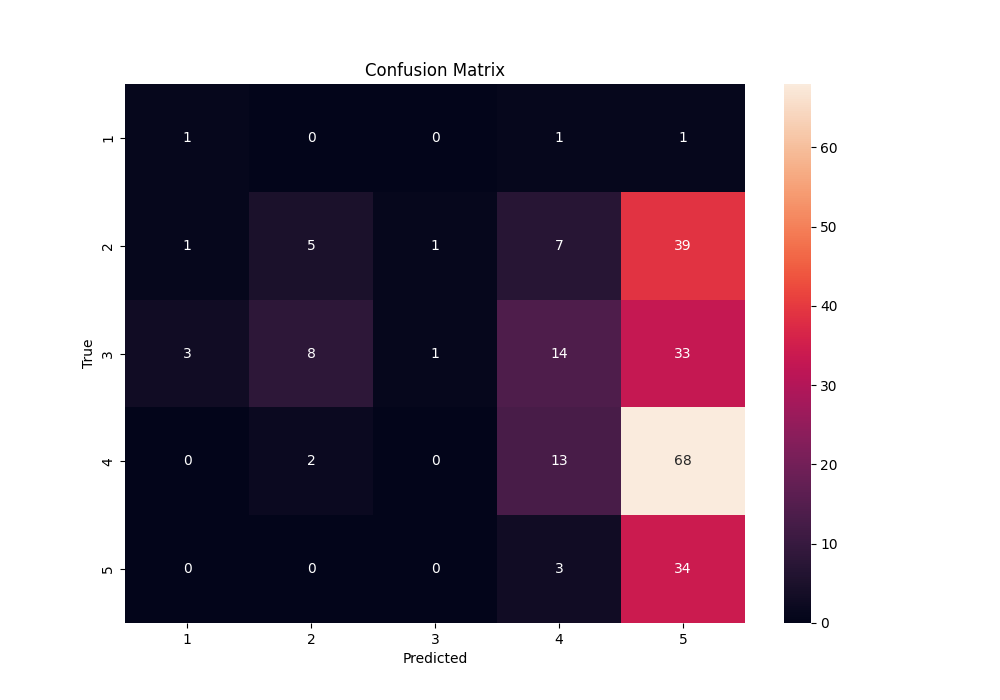}
        \quad %
        \includegraphics[width=0.48\textwidth]{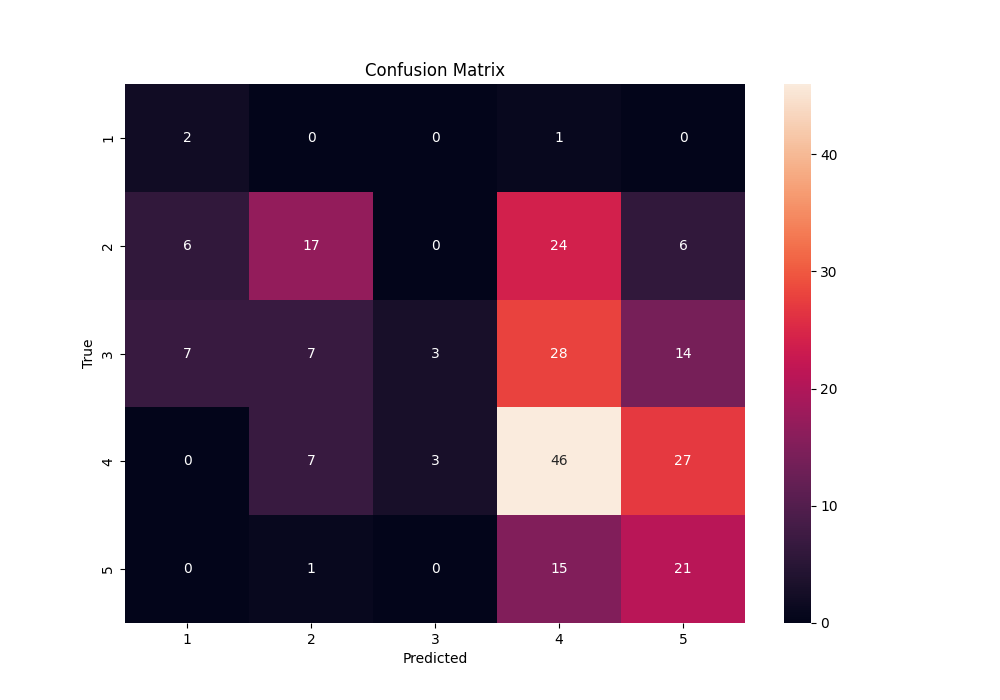}
        \subcaption{GPT-4-turbo zero-shot vanilla (left), GPT-4-turbo few-shot vanilla (right)}
    \end{minipage}
    
    \vspace{-0.2em}
    
    \begin{minipage}{\textwidth}
        \centering
        \includegraphics[width=0.48\textwidth]{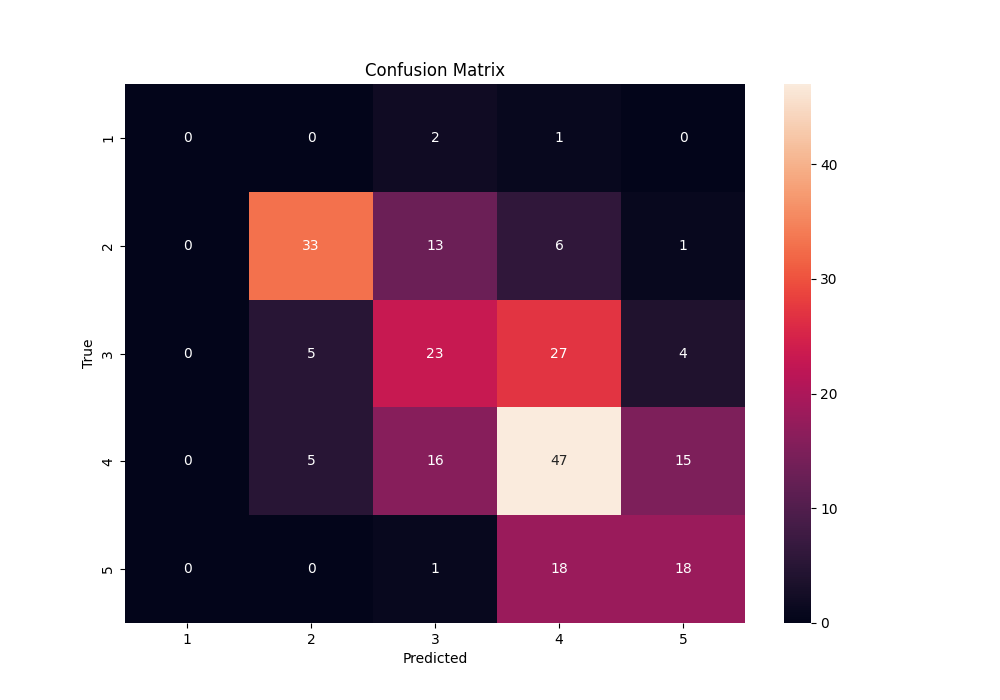}
        \quad %
        \includegraphics[width=0.48\textwidth]{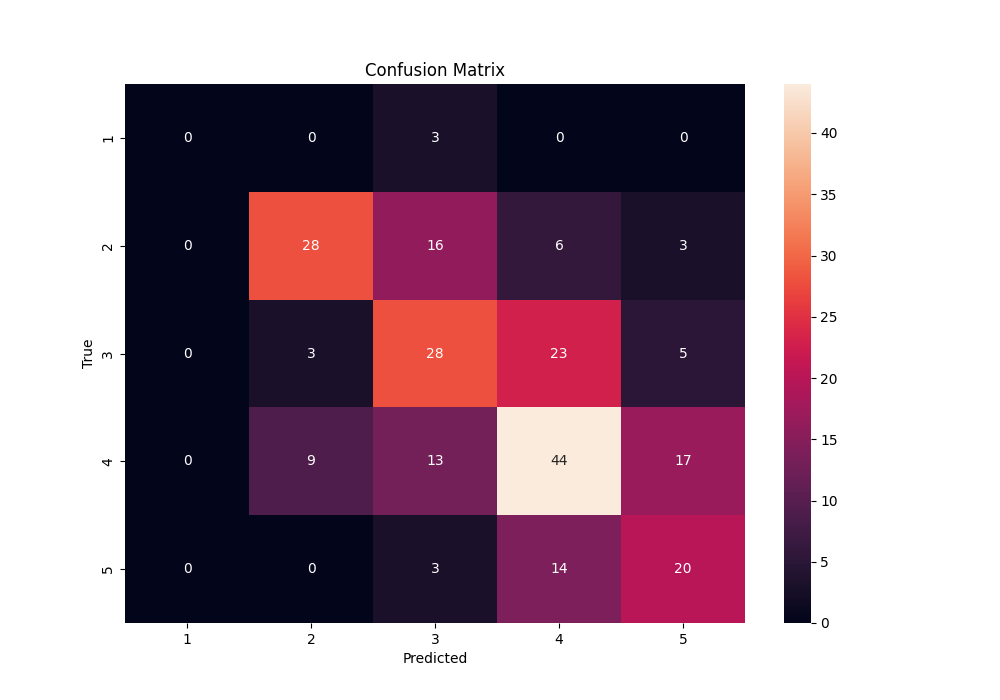}
        \subcaption{Mistral-7B-Instruct (left),  Llama-2-7B-chat (right)}
    \end{minipage}
    
    \vspace{-0.2em}

    \begin{minipage}{\textwidth}
        \centering
        \includegraphics[width=0.48\textwidth]{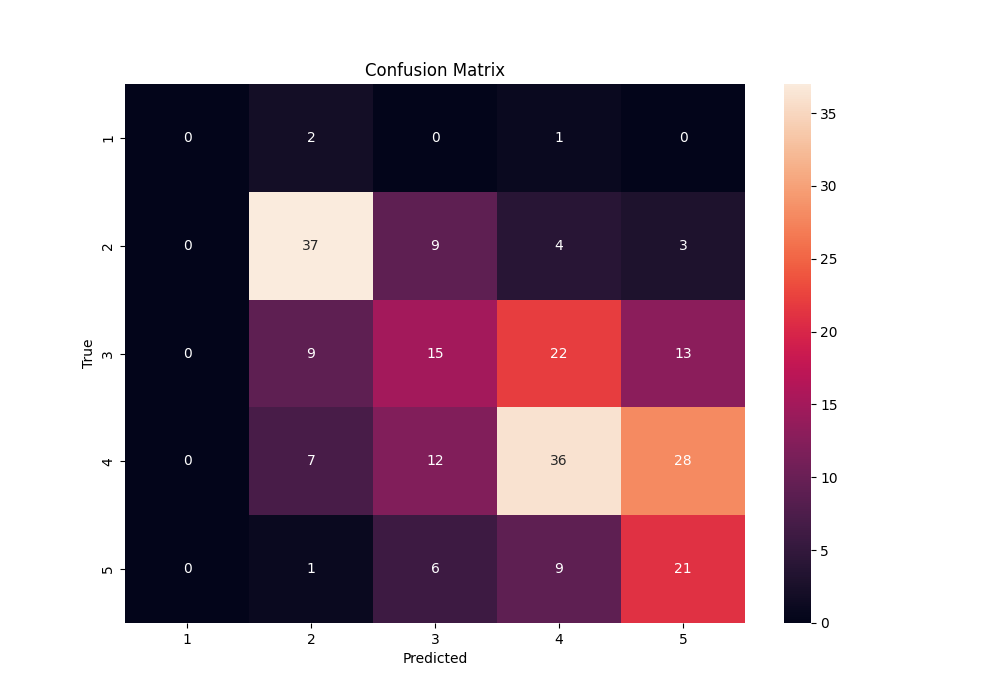}
        \quad %
        \includegraphics[width=0.48\textwidth]{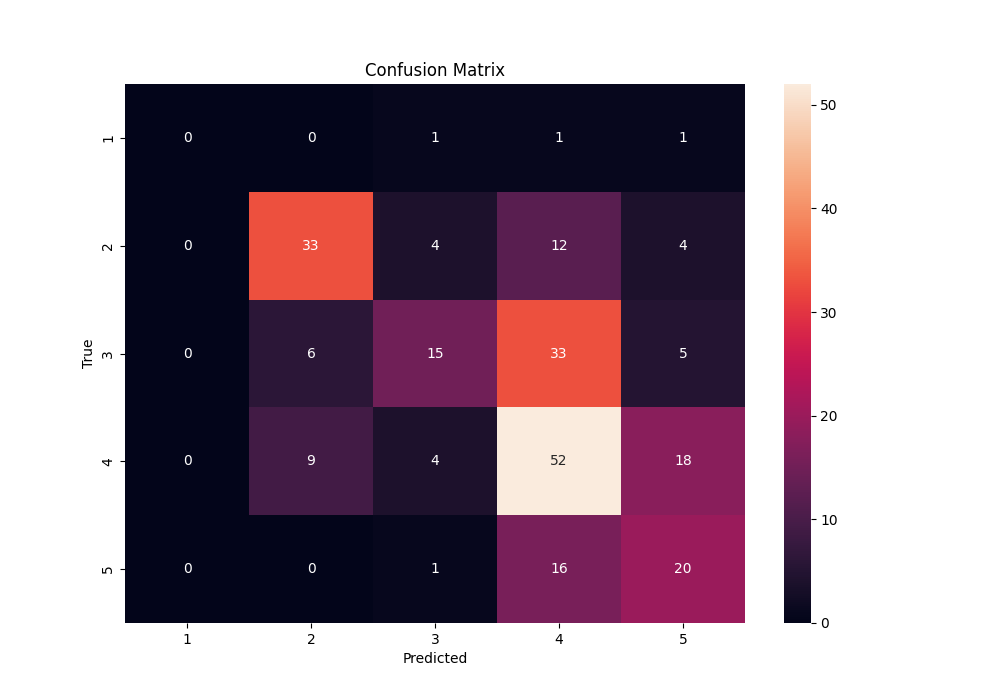}
        \subcaption{Flan-t5-base (left), TinyLlama-1.1B-chat (right)}
    \end{minipage}

\caption{confusion matrices for salience prediction across different models.}
\label{fig:confusion_matrix}
\end{figure*}

\begin{prompt}[title={\thetcbcounter{} Prompt for GPT-4 summary expansion.},label=gpt4:vanilla:elab]
    \promptsubsection{article}\highlight{article}
    
    \promptsubsection{short summary}\highlight{tl;dr}

    Produce an elaboration of the short summary by including relevant details from the article within a word count range of 230 to 250 words. Strive for conciseness and clarity in delivering a comprehensive expansion within the specified word limit. The response MUST NOT exceed 250 words at any cost. Produce outputs less than 250 words.

    \promptsubsection{elaboration}
\end{prompt}

\begin{prompt}[title={\thetcbcounter{} Prompt for GPT-4-Corrupted expansion (Step 1).},label=gpt4:elab:step1]
\promptsubsection{article} \highlight{article}

\promptsubsection{short summary} \highlight{tl;dr}

Read the article and the short summary. Provide a list of all the important topics from the short summary and related to it which are spoken about in the article. Your response should be a comma separated list.

\promptsubsection{response}
\end{prompt}

\begin{prompt}[title={\thetcbcounter{} Prompt for GPT-4-Corrupted expansion (Step 2).},label=gpt4:elab:step2]
    \promptsubsection{article}\highlight{article}

    \promptsubsection{irrelevant topic}\highlight{irrelevant topic}

    In 230 to 250 words, produce an elaboration of the article by omitting as many topics included or related to the 'irrelavant topic' field as possible. Your response MUST be strictly more than 230 words and under 250 words. Remember, you MUST produce ATLEAST 230 word count responses.

    \promptsubsection{response}
    
\end{prompt}

\begin{prompt}[title={\thetcbcounter{} Prompt for improving the expansion style while imposing minimal content changes.},label=gpt4:elab:style]
    \promptsubsection{paragraph}\highlight{original expanded summary}

    Make minor alterations to the paragraph above such that its narrative style is similar to a usual summary. Do not use very flowery language and stick to the contents in the paragraph ONLY. Your response should NOT include any new content. Your response should be over 230 words but not exceed 250 words. Remember, do not produce responses below 230 words. Don't start the sentences like the 'article talks about this' or 'the article sheds light on..'. Remember, you MUST produce ATLEAST 230 word count responses.

    \promptsubsection{response}:
\end{prompt}

\section{Prompts for Expanded Summary Evaluation}\label{app:summary-eval-prompts}

\begin{prompt}[title={\thetcbcounter{} Prompt for identifying questions that are unanswered in the article.},label=remove:unans]
    \promptsubsection{article}\highlight{article}

    Which sentences from the article completely answer the question \highlight{question} Include only the relevant sentences extracted from the article that are answers to the question and NOT just vaguely related to the topic introduced in the question. Be concise. Your response should not exceed 3 lines. If the article doesn't provide a SPECIFIC answer to the question, respond with 'No Answer'.

    \promptsubsection{response}

\end{prompt}

\begin{prompt}[title={\thetcbcounter{} Prompt for finding questions that an expanded summary answers.},label=ans:questions]
    \promptsubsection{article}\highlight{expanded summary}
    
    \promptsubsection{questions}\highlight{list of questions}

    Read the above article and find the questions from the 'questions' list provided above which are answered in the article. Your response should be a comma separated list of only questions that are completely or partially answered by the article.
    
    \promptsubsection{response}
\end{prompt}

\begin{table}[b!]
\centering
\small
\begin{tabular}{ll}
\toprule
\textbf{Question Type} & \textbf{PMI} \\
\midrule
CONSEQUENCE & 0.413 \\
EXAMPLE & 0.279 \\
PROCEDURAL & 0.146 \\
COMPARISON & 0.127 \\
CAUSE & 0.019 \\
EXTENT & -0.060 \\
VERIFICATION & -0.112 \\
DISJUNCTIVE & -0.401 \\
CONCEPT & -0.451 \\
JUDGMENTAL & -0.648 \\
\bottomrule
\end{tabular}
\caption{Question types ranked by PMI with high salience scores.}
\label{tab:q_typ_pmi}
\end{table}

\begin{figure*}[ht]
    \centering
    \includegraphics[width=\textwidth,height=\textheight,keepaspectratio]{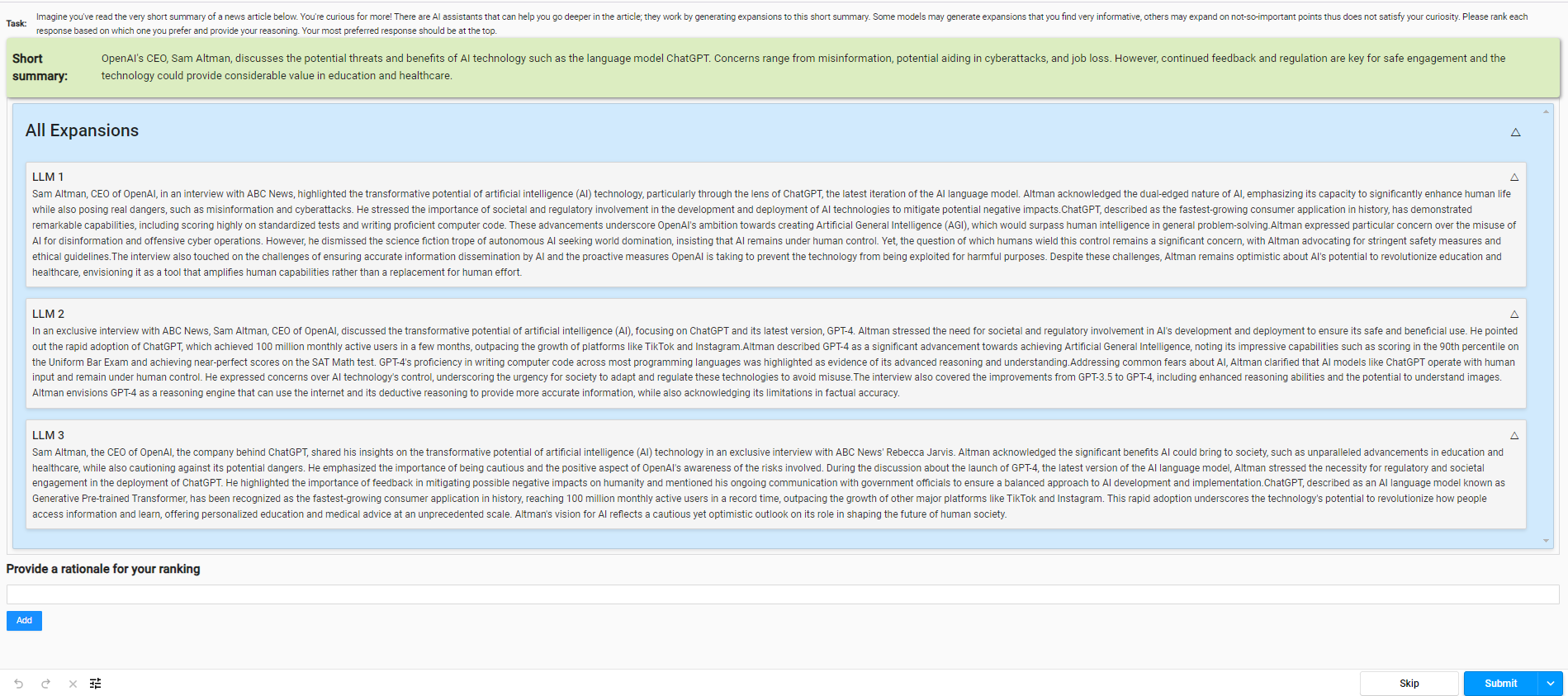}
\caption{Annotation interface for the summary expansion task; the three candidates are ordered via drag-and-drop.}
\label{fig:annotation_interface}
\end{figure*}

\begin{table*}
\small
\centering    
\begin{tabular}{p{15cm}}
\toprule
\textbf{article}: [1] The desperate actions by governments, regulatory authorities, and banks in both the US and Europe have not only failed to stem the growing financial crisis but in some ways are making it worse.[2] In the US, following the failure of the Silvergate bank, Silicon Valley Bank and Signature over the past two weeks, the latter two recording the second- and third-largest banking failures in US history respectively, attention has turned to the travails of the First Republic Bank with growing concerns that it could be the next to go.[3] Last week, a consortium of 11 major banks, under the leadership of JPMorgan Chase CEO Jamie Dimon, with the collaboration of Treasury Secretary Janet Yellen, deposited \$30 billion with the struggling bank.[4] It was hoped this show of confidence would stop the outflow of depositors' money, ease the pressure on its share price and stabilise it.\newline
\textbf{questions}:\\
1) Who initiated the act of depositing \$30 billion into the struggling First Republic Bank? \textbf{Salience = 2}\\
2) Why was it thought that this deposit would stem the outflow of depositors' money? \textbf{Salience = 5}\\
3) What role did Treasury Secretary Janet Yellen play in this financial effort? \textbf{Salience = 0}\\
\midrule
\textbf{article}: [5] In just a few days, the operation has been revealed as a complete failure.[6] While the outflows are reported to have slowed somewhat, First Republic has lost \$70 billion out of the total of \$176 billion it held at the start of the year.[7] And despite the injection of cash, the company's shares have continued to plummet.\newline
\textbf{questions}:\\
1) How has the injection of cash affected the overall financial health of the company? \textbf{Salience = 3}\\
2) What are the strategies that the company intends to use to stabilize its shares amid the injection of cash? \textbf{Salience = 4}\\
\midrule
\textbf{article}: [8] Its share price has fallen by 90 percent since the beginning of the month, closing 47 percent down yesterday.[9] Long-term bonds that mature in 2046 were trading at 55 cents on the dollar, down from 75 cents in early March.[10] First Republic took another hit before trading opened yesterday, when the ratings agency S\&P Global downgraded its credit rating for the second time in a week.\newline
\textbf{questions}:\\
1) What was First Republic's credit rating prior to these two downgrades by S\&P Global? \textbf{Salience = 3}\\
2) Why did the ratings agency S\&P Global downgrade First Republic's credit rating for the second time in a week? \textbf{Salience = 5}\\
\bottomrule
\end{tabular}
\caption{An example of an article with potential questions it evokes and their corresponding human-annotated salience scores.}
\label{tab:salience-example}

\end{table*}

\begin{table*}
\small
\centering
\begin{tabular}{p{15cm}}
\toprule
\textbf{TL;DR}: The IMF reached a staff-level agreement to offer Ukraine a \$15.6bn financial package, marking the first loan to a country at war. The two-phase agreement involves enhancing fiscal and financial stability, followed by more extensive reforms. The loan follows recent amendments allowing the IMF to fund countries facing 'exceptionally high uncertainty'.\newline

\textbf{best expansion}: The International Monetary Fund (IMF) has agreed to provide Ukraine with a significant financial package worth \$15.6 billion, a groundbreaking move as it marks the first instance of the organization offering a loan to a nation currently engaged in war. This decision comes after the IMF recently amended its policies to permit funding under conditions of "exceptionally high uncertainty," enabling support for countries grappling with extraordinary challenges such as war or severe natural disasters.
The financial aid is structured into a two-phase program spanning four years. Initially, Ukraine will focus on enhancing fiscal, external, price, and financial stability, including the cessation of monetary financing, over a period of 12-18 months. The subsequent phase will pivot towards broader reforms aimed at securing macroeconomic stability, supporting recovery and early reconstruction efforts, and fostering resilience and higher long-term growth. These reforms are particularly geared towards Ukraine's aspirations for European Union accession, with an emphasis on reverting to pre-war policy frameworks like a flexible exchange rate and inflation targeting regime.
This loan package is not only the largest the IMF has proposed for a country amidst conflict but also a testament to the international community's support for Ukraine during its ongoing war with Russia. The IMF's decision reflects a recognition of the severe economic impact of the conflict on Ukraine, including a 30\% contraction in economic activity in 2022, significant destruction of capital stock, and rising poverty levels. By securing this financial assistance, Ukraine aims to stabilize its economy, mobilize further support from international partners, and lay the groundwork for post-war reconstruction and recovery. \newline

\textbf{good expansion}: The International Monetary Fund (IMF) has agreed to provide Ukraine with a \$15.6 billion financial package, a significant move as it's the first loan offered to a country engaged in war. This agreement, structured in two phases, aims to first enhance Ukraine's fiscal and financial stability, with a subsequent focus on broader reforms. This decision comes after recent changes in the IMF's policies, now allowing funding for countries experiencing 'exceptionally high uncertainty'. The journey to this agreement began last June when Ukraine first initiated discussions with the IMF, leading to a final approval by the IMF executive board on March 20, after an initial round of talks failed and a second was launched in August.
Meanwhile, Pakistan faces challenges in securing a vital IMF bailout package. The government revealed that the IMF is seeking financial support for Pakistan from 'friendly' nations before proceeding with a \$6.5 billion bailout program. This stance was reiterated with a slight variation in the required amount, mentioning a \$6.1 billion tranche.
Criticism has been directed at the IMF's approach to assisting poor countries. Nobel Prize-winning economist Joseph Stiglitz, in his 2002 book "Globalisation and Its Discontents," criticized the IMF's economic reform conditions. He argued that these often counterproductive measures have had devastating effects on the populations of the target countries, highlighting a broader debate on the IMF's role and policies in global financial stability.\newline

\textbf{worst expansion}: The International Monetary Fund (IMF) has updated its policies to support countries experiencing "exceptionally high uncertainty," aiming to assist those affected by crises like wars or natural disasters. This policy revision, revealed on March 17, diverges from the IMF's earlier approach that restricted its capacity to offer non-emergency loans under conditions of significant uncertainties. The modification in the financing assurances policy intends to cover scenarios where external shocks, beyond a country's economic policy influence, require financial aid.

The IMF has been instrumental in aiding countries during financial crises by providing loans to address challenges such as foreign debt payments and reducing foreign exchange reserves. Funding for these loans is sourced from the capital subscriptions or quotas contributed by member countries, reflecting their economic standing globally. With 190 member countries, the United States is the largest contributor to the IMF.
Before approving a loan, the IMF conducts discussions with the applicant country to assess its financial situation and requirements. The country typically needs to agree to undertake certain economic policy measures. Following an agreement, the IMF's executive board reviews and approves the loan, which is then disbursed in phases. Throughout this period, the IMF monitors the country's adherence to the agreed economic policies. This procedure highlights the IMF's commitment to promoting economic stability and reform among its member countries facing financial challenges.\\
\bottomrule
\end{tabular}
\caption{An example of GPT-4, Flan-T5 and GPT-4-Corrupted expansions from best to worst.}
\label{tab:elab:full:example}
\end{table*}

\section{Question Type vs Salience}\label{app:qtype}
We computed PMI (pointwise mutual information) between question type and salience scores to see if they relate. We divided salience into three levels: low, mid, high. We classified questions into 13 types defined in \cite{cao2021controllable}. The ranked PMI by the highest salience level can be found under Table \ref{tab:q_typ_pmi}.

\section{Compute}
We used 3 NVIDIA A40 (48GB) and 1 A100 (80GB) for fine-tuning models. For A40, each training took less than 30 minutes. For A100, each training took under 10 minutes. The training process for all models finished within 4 hours.

\section{License and Copyright}
We will release our annotations under the Creative Commons CC-BY license; the original texts (DCQA \cite{ko-etal-2022-discourse}, TED-Q \cite{westera-etal-2020-ted} and DiverseSumm \cite{huang2023embrace}) come with their original licenses.

\end{document}